\documentclass[runningheads]{llncs}
\usepackage{graphicx}
\usepackage{epstopdf}
\usepackage{times}
\usepackage{epsfig}
\usepackage{graphicx}
\usepackage{amsmath}
\usepackage{amssymb}
\usepackage{algorithm}
\usepackage{algpseudocode}
\usepackage{amsmath}
\usepackage{multirow}
\usepackage{subfigure}
\usepackage{booktabs}
\usepackage{color}
\usepackage{booktabs}
\usepackage{threeparttable}
\usepackage{footnote}
\usepackage{hyperref}
\begin{document}
\title{Multi-modal Graph Learning for Disease Prediction}
%
%
\author{Shuai Zheng\inst{1,2} \and
Zhenfeng Zhu\inst{1,2} \and
Zhizhe Liu\inst{1,2} \and
Zhenyu Guo\inst{1,2} \and
Yang Liu\inst{1,2,3} \and
Yao Zhao\inst{1,2}}
%
%
\institute{Institute of Information Science, Beijing Jiaotong University, Beijing, China  \and
Beijing Key Laboratory of Advanced Information Science and Network Technology, Beijing, China\\
\and
School of Information Science and Engineering, Hebei North University, Zhangjiakou, China \\ \email{\{zs1997,zhfzhu,zhzliu,zhyguo,19112005,yzhao\}@bjtu.edu.cn}  }
\maketitle              
\begin{abstract}
Benefiting from the powerful expressive capability of graphs, graph-based approaches have achieved impressive performance in various biomedical applications. 
Most existing methods tend to define the adjacency matrix among samples manually based on meta-features, and then obtain the node embeddings for downstream tasks by Graph Representation Learning (GRL).
However, it is not easy for these approaches to generalize to unseen samples. Meanwhile, the complex correlation between modalities is also ignored. As a result, these factors inevitably yield the inadequacy of providing valid information about the patient's condition for a reliable diagnosis.
In this paper, we propose an end-to-end \underline{M}ulti-\underline{m}odal \underline{G}raph \underline{L}earning framework (MMGL) for disease prediction. 
To effectively exploit the rich information across multi-modality associated to diseases, a modal-attentional multi-modal fusion is proposed to integrate the features of each modality by leveraging the correlation and complementarity between the modalities. 
Furthermore, instead of defining the adjacency matrix manually as existing methods, the latent graph structure can be captured through a novel way of adaptive graph learning. It could be jointly optimized with the prediction model, thus revealing the intrinsic connections among samples. 
Unlike the previous transductive methods, our model is also applicable to the scenario of inductive learning for those unseen data.
An extensive group of experiments on two disease prediction problems is then carefully designed and presented, demonstrating that MMGL obtains more favorable performances. In addition, we also visualize and analyze the learned graph structure to provide more reliable decision support for doctor in real medical applications and inspiration for disease research.

\end{abstract}
\vspace{-1cm}
\section{Introduction}
\vspace{-0.2cm}
A large amount of relational data exists in the biomedical field. As a general description of relational data, graphs are facilitated to model a variety of biomedical scenarios \cite{survey}. Recently, inspired by the excellent performance of graph-based methods in machine learning, particularly graph convolutional networks (GCNs) \cite{GCN,GAT,Graphsage}, they have also been applied to handle relational data in various Computer Aided Diagnosis (CADx) tasks, such as Alzheimer prediction \cite{TADPOLE}, Autism prediction \cite{ABIDE}, and cancer prognosis prediction \cite{hybrid}. 

\begin{figure}[t]	
	\centering
	\includegraphics[width=4in]{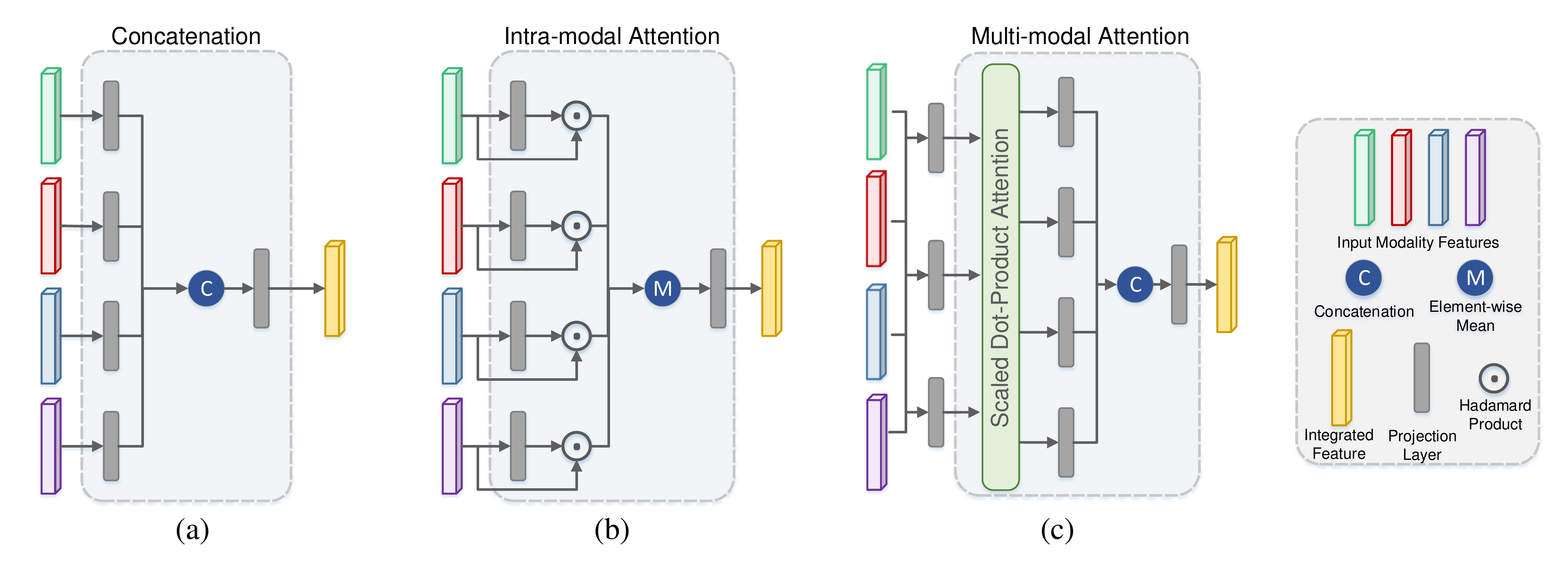}
	\vspace{-0.5cm}
	\caption{Architecture of three typical types of multi-modal feature fusion.
		(a) Concatenation fusion,
		(b) Weighted fusion based on intra-modal attention mechanism.
		(c) Our modal-attentional feature fusion.
		 The (a) and (b) fusions have only one interactive operation for different modal features, which proceed at the end of a module. In contrast, our module has more interactive operations for different modal features. 
	}
	\vspace{-0.5cm}
	\label{fusion}
\end{figure}

Most existing methods try to construct the patient relationship graphs from existing features through pre-defined similarity measures, then apply GCNs to aggregate patient features over local neighborhoods to give the prediction results.
These methods can be broadly classified into two categories: single-graph-based methods and multi-graph based methods. 
Parisot et al. \cite{popGCN} proposed to compute patient similarities from a set of meta-features such as age and sex, thus constructing the adjacency matrix to apply GCNs. 
\cite{InceptionGCN,multi-hop} employed the same graph construction rules and initially explored the effect of graph structure on performance of the disease prediction through setting different neighborhood sizes.
These methods simply combine the imaging and non-imaging modalities through graph construction and GCNs. However, they fail to effectively mine the intrinsic information of each modality. 
Thus, several recent works have proposed to construct multiple graphs in parallel, in which each graph is built from different modality, then execute the integration of the embeddings learned from different graphs for the prediction.
As shown in Fig.\ref{fusion}(a),~\cite{MGNN} concatenated the node embeddings directly, and both~\cite{selfGCN} and~\cite{recurrent} adopted attention-based fusion mechanism like Fig.\ref{fusion}(b).

Although the above methods have achieved remarkable performance, three key issues remain to be further considered with respect to the application of GCNs in disease prediction tasks, and even in some other biomedical tasks:

\textbf{\emph{(1) Insufficient inter-modal relationship mining.}} Each modality provides different information for the diagnosis of a disease, which is both complementary and potentially redundant. However, concatenation \cite{LGL,MGNN,huang} or intra-modal attention mechanism \cite{selfGCN,recurrent} adopted in previous studies are hard to capture the latent \textit{inter-modal correlation}, which may lead to the learned features are biased towards a single modality. To solve this issue, we propose a modal-attentional multi-modal fusion to mine the intrinsic relevance between modalities while preserving the individuality of each modality.

\textbf{\emph{(2) Defining the graph adjacency matrix manually.}} Existing single-graph-based \cite{popGCN,InceptionGCN,yang} and multi-graph-based methods \cite{LSTMGCN,MGNN,recurrent} both construct the graph through hand-designed similarity measures, which requires a careful tuning and is thus difficult to generalize to downstream tasks. A better approach is to learn a graph by  the means of end-to-end, but less focus has been put on the graph structure learning\cite{adaptive,graph_learning}, especially in the field of medicine\cite{LGL}. Meanwhile, sigmoid-based graph learning mechanism in \cite{LGL} is prone to causing gradient disappearance, which makes model training unstable. Thus, we propose a learning-based adaptive approach for graph learning to learn the graph structure dynamically. In fact, it provides a more feasible way for downstream tasks and reveals the latent connections among samples.


\textbf{\emph{(3) Hard applicable to inductive learning.}} 
For the approaches based on spectral graph convolution like~\cite{huang,InceptionGCN,popGCN}, it's hard for them to generalize to unseen samples. Besides, to accommodate inductive learning, it is also essential but cumbersome for multi-graph-based methods \cite{LSTMGCN,MGNN,recurrent} to measure relationship of unseen samples on each graph. Unlike these approaches, our MMGL can be flexibly extended to the scenario of inductive learning.

In this paper, we propose a novel \underline{M}ulti-\underline{m}odal \underline{G}raph \underline{L}earning framework (MMGL) for disease prediction, and the main contributions can be highlighted in the following aspects:
\begin{itemize}
	\item As a flexible modular inductive learning framework, the proposed MMGL provides some substantial improvements and inspirations for the application of GCN in disease prediction tasks.

	\item Considering the correlation and complementarity between modalities, we propose a modal-attentional feature fusion approach (MaFF) that exploits the inter-modal relevance to integrate the multi-modal features.

	\item To reveal the intrinsic connections among samples, a novel adaptive graph learning mechanism (AGL) is proposed to achieves learnable graph construction, thus obtaining the latent robust graph for downstream tasks.

	\item The comparable even significant improvement compared to the state-of-the-art methods indicates the advantages of our MMGL in terms of disease prediction tasks.
\end{itemize}
\vspace{-0.7cm}
\section{Methodology}
\vspace{-0.3cm}
\subsection{Problem Formulation}
\vspace{-0.2cm}
Let $X= \left [x_1, x_2, \cdots, x_N \right ] \in \mathbb{R}^{d_{in} \times N}$ denote the $d_{in}$-dimensional multi-modal features of $N$ patients and $Y=\left [y_1, y_2, \cdots, y_N\right ]$ is the corresponding labels. Each patient is represented by $M$ modalities, so the multi-modal features also can denote as $X=\{X^1, X^2, \cdots, X^M\}$ and $X^m = \left [x^m_1, x^m_2, \cdots, x^m_N \right ]\in \mathbb{R}^{d_m \times N}$, where $x^m_i$ represents the $d_m$-dimensional features of the $m$-th modality of $x_i$ and $d_{in} = \sum_{m=1}^{M}d_m$. Given the multi-modal features $X$, the issue we consider in this paper for multi-modal graph learning is to achieve an optimal latent graph structure inference based on multi-modal feature fusion considering inter-modal correlation, thus providing reliable graph support to GCNs for disease prediction or the other biomedical tasks.

\begin{figure}[t]	
	\centering
	\includegraphics[width=4.3in]{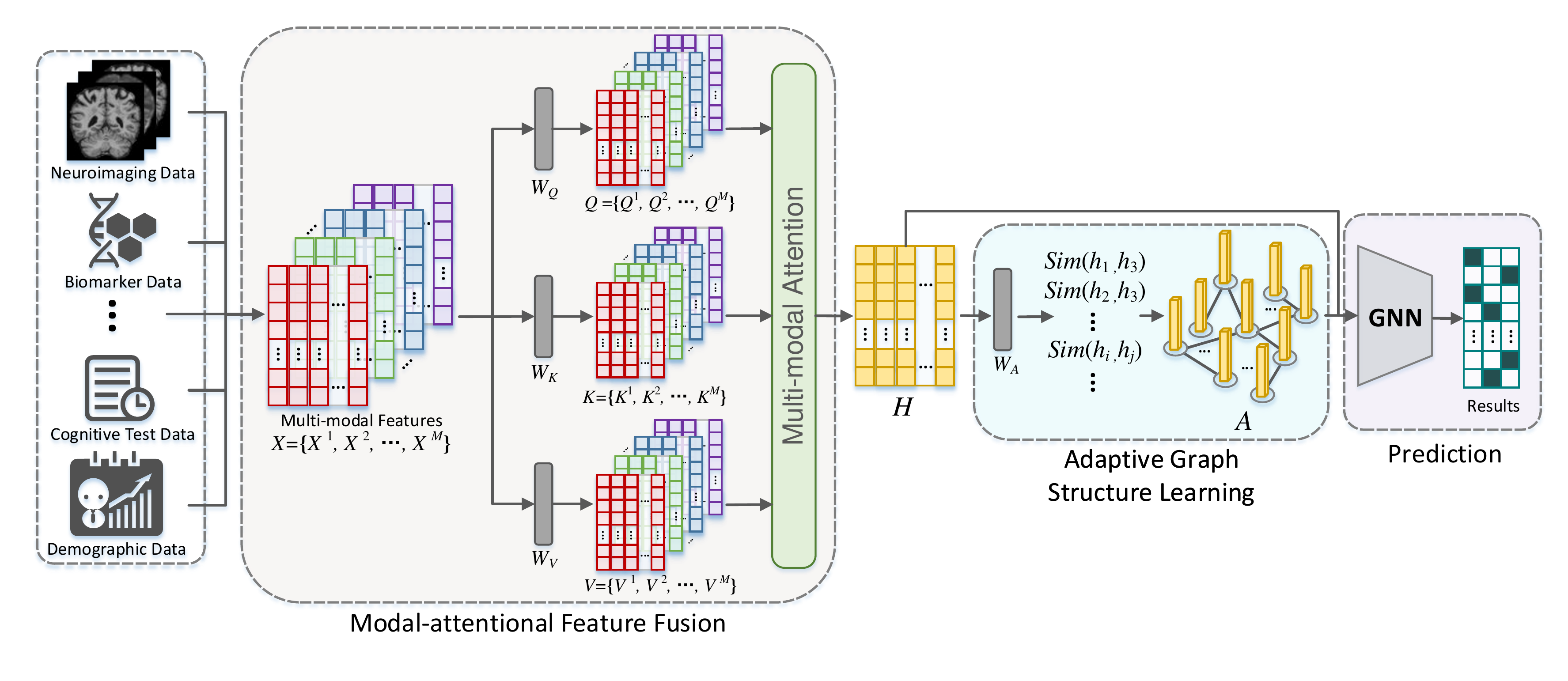}
	\vspace{-0.5cm}
	\caption{Architecture overview of our MMGL. MMGL consists of three modules: Modal-attentional feature fusion module, adaptive graph learning module, and prediction module.
	}
	\vspace{-0.5cm}
	\label{Overview}
\end{figure}

As illustrated in Fig.~\ref{Overview}, the overall framework of MMGL consists of three modules. In feature fusion phase, the multi-modal features $X$ is integrated into a fused single feature matrix $H=\left [h_1,h_2,\cdots,h_N\right ]\in \mathbb{R}^{d \times N}$ by modal-attentional feature fusion module (MaFF). Then, in graph learning phase, the adjacency matrix $A \in \mathbb{R}^{N \times N}$ characterizing the latent graph structure is captured through a novel adaptive graph learning mechanism (AGL) based on the multi-modal modal-fused features. Finally, according to the learned adjacency matrix $A$ and fused feature matrix $H$, the disease prediction results can be obtained by GCN. 

\vspace{-0.4cm}
\subsection{Modal-attentional Feature Fusion}
\vspace{-0.2cm}
In a real diagnostic scenario, medical experts always need to analyze various multi-modal data of the patient to make a reliable decision, since the single-modal data lacks of providing enough information for an accurate diagnosis. Similarly, reliable Computer Aided Diagnosis (CADx) also requires to be capable of leveraging the complementarity of multi-modal data~\cite{CAD}. To capture inter-modal relevance and complementarity for efficient information integration, we propose a transformer-style~\cite{transformer} modal-attentional feature fusion module named \textbf{MaFF}, which fuses the features of each modality while preserving rich information.

Given the query vectors $Q^m$ of current modality and key-value pairs of the others (i.e., $K$-$V$), the inter-modal attention score can be calculated. In practice, considering the problems of space-efficiency and parallelization, the scaled dot-product is chosen as attention function~\cite{transformer}.
Since each modal feature has different dimension, as shown in Fig.~\ref{fusion}(c), we first use the projection matrices (i.e., $W_Q$, $W_K$, $W_V$) with fixed dimension $d_f$ to translate $X$ to $Q$, $K$, and $V$, which facilitates the computation of attention scores in the same dimensional space. For a specific patient $x$, we can obtain the corresponding query matrix $q=\{q^m\}^M_{m=1}$, key matrix $k=\{k^m\}^M_{m=1}$, and value matrix $v=\{v^m\}^M_{m=1}$. Then, the inter-modal attention score map $P$ for $x$ is computed as:
\begin{equation}
P_{ij} = \frac{\mathrm{exp}[(q^i)^\top\cdot k^j/\tau]}{\sum_{i=1}^{M}\mathrm{exp}[(q^i)^\top\cdot k^j/\tau]}
\label{attention}
\end{equation}
where $P_{ij}$ denotes the attention score between modality $i$ and $j$. $\tau$ is the scaling factor to control the hardness of attention, which is set to $\sqrt{d_f}$ like in ~\cite{transformer}. After that, we perform cross-modal aggregation of the value vectors of each modality based on the calculated inter-modal correlations $P$, thus the integrated feature $h$ of patient $x$ could be computed as:
\begin{equation}
h = W_h^\top \cdot Concat(\hat{v}^1,\hat{v}^2,\cdots,\hat{v}^M),~\mathrm{where}~\hat{v}^m=W_m^\top\cdot(v^m+\sum_{j=1}^{M}P_{mj}~v^j)
\label{value}
\end{equation}
where $W_h$ and $W_m$ are projection layers. Besides, the implementation of residual connection between $\sum_{j=1}^{M}P_{m,j}~v^j$ and the initial value vector $v^m$ can effectively avoid the gradient vanishing problem during training process. 

Compared to the fusion module in~\cite{selfGCN,LGL,MGNN}, the modal-attentional fusion puts concerns on inter-modal correlation through a multi-modal attention mechanism. As a consequence, it tend to optimally combine the complementary information from different modalities. It's worth noting that the multi-modal attention map $P$ is patient-specific, which is also applicable to inductive learning.
And we could obtain global-level inter-modal correlation map by averaging $P$ over all patients. Besides,the proposed multi-modal attention mechanism can also be scalable to multi-head version easily.
\vspace{-0.4cm}
\subsection{Adaptive Graph Structure Learning}
\vspace{-0.2cm}
Based on existing graph structures, GCNs in~\cite{GCN,GAT,ChebNet} learn node representations for downstream tasks through neighborhood aggregation or spectral convolution. However, it's no trivial to obtain the available graph for some specific tasks in biomedical field. Therefore, the graph learning problem often needs to be considered for applying GCNs in biomedical tasks. 

For graph learning, it is usually modeled as two kinds of forms: \textbf{(i)} learning a joint discrete probability distribution on the edges of the graph~\cite{discrete}, \textbf{(ii)} learning a similarity metric of nodes. Since the former is non-differentiable and hard applicable to inductive learning, we take the graph learning problem in consideration from the perspective of similarity metric learning of nodes. 
Some previous methods have adopted radial basis function (RBF) kernel~\cite{recurrent,multi-hop}, cosine similarity~\cite{huang}, or threshold-based metric(for discrete feature)~\cite{popGCN,InceptionGCN} as the similarity metric. However, these approaches still require careful manual tuning to construct a meaningful graph structure for downstream GCNs. Therefore, as illustrated in Fig~\ref{Overview}, we propose a simple but effective learnable metric function, which could be jointly optimized with the downstream GCNs:

\begin{equation}
\hat{A}_{ij}= Sim(h_i, h_j) = \cos( W_A^\top h_i,~W_A^\top h_j)
\label{graph}
\end{equation}
where $W_A$ is a learnable weight matrix and $\hat{A}_{ij}$ is computed as weighted cosine similarity between patient $i$ and $j$. Since there are few uni-directional effects between patients except for epidemics, thus the learned adjacency matrix $\hat{A}$ is symmetric that is also in accordance with the expectations of realistic population graph of patients. 

Commonly, a realistic adjacency matrix is usually non-negative and sparse. However, since $\hat{A}$ is a fully connected graph that is computationally expensive and the element $\hat{A}_{ij}$ is ranging in [-1,1], we capture a non-negative sparse graph $A$ from $\hat{A}$ by applying the ReLU function, i.e., setting negative value elements in $\hat{A}$ to zero. Finally, the latent graph $A$ is obtained for downstream tasks.

\vspace{-0.4cm}
\subsection{Model Optimization}
\vspace{-0.2cm}
\textbf{Graph Regularization} Due to the sensitivity of GCNs to the graph structure, graph learning has a significant impact on the performance of GCN in downstream tasks. Furthermore, the constraint on the sparsity, connectivity, and smoothness of the learned graph is also important for adaptive graph learning~\cite{IDGL}. Here, the Dirichlet energy is used to measure the smoothness of a set of graph signals $\{h_1, h_2, \cdots, h_N \}$:
\vspace{-0.2cm}
\begin{equation}
\mathcal{L}_{smooth} =\frac{1}{2 N^{2}} \sum_{i, j=1}^{N} A_{i j}\left\|h_{i}-h_{j}\right\|^{2}
\label{smoothness}
\end{equation}

It can be seen from Eq.~\ref{smoothness} that the smaller the distance between $h_i$ and $h_j$, the larger $A_{ij}$ will be. Hence, the smooth loss $\mathcal{L}_{smooth}$ is intended to making connections between similar nodes, which means to enforce smoothness of the graph signals on the learned graph $A$.Essentially, $L_{smooth}$ simultaneously serves to control the sparsity of A~\cite{graph_learning}. However, only utilizing $L_{smooth}$ may lead to the trivial solution (i.e., $A = 0$). To avoid it, two additional regularization terms following~\cite{graph_learning} are imposed on $A$:
\begin{equation}
\mathcal{L}_{con} = \frac{-1}{n} \mathbf{1}^{\top} \log (A \mathbf{1})
~~~\mathrm{and}~~~
\mathcal{L}_{r} = \frac{1}{n^{2}}\|A\|_{F}^{2}
\label{c_and_s}
\end{equation}
where the first term $\mathcal{L}_{con}$ uses logarithmic barrier to control the connectivity of $A$, and the second term $\mathcal{L}_{r}$ is a regularization term to avoid the excessive sparseness caused by $\mathcal{L}_{smooth}$. The total graph regularization loss is defined as $\mathcal{L}_{g} = \mathcal{L}_{smooth} + \alpha \mathcal{L}_{con} + \beta \mathcal{L}_{r}$.

\subsubsection{Loss Function} Based on the integrated feature matrix $H$ and learned sparse graph $A$, we use GNNs to give the predicted results $\hat{Y} = GCN(H,A)$ of the patients. 
Unlike~\cite{LGL,huang} which optimize the graph structure directly based on task-aware prediction loss, we use a joint loss function to guide the optimization of all three modules of MMGL simultaneously:
\begin{equation}
\mathcal{L} = \mathcal{L}_{t}(Y, \hat{Y})  + \lambda \mathcal{L}_{g}(H,A)
\label{total_loss}
\end{equation}
where $\mathcal{L}_{t}$ and $\mathcal{L}_{g}$ denote the task-aware loss and the graph regularization loss respectively, $\lambda$, $\alpha$, and $\beta$ are hyper-parameters to balance the three loss terms. For the disease prediction task treated as classification problems, $\mathcal{L}_{t}$ is set to cross-entropy loss.
\vspace{-0.4cm}
\section{Experimental Results and Analysis}
\vspace{-0.2cm}
In this section, we evaluate the performance of MMGL on two biomedical datasets. We first detail our experimental protocol, and then present the comparison results of MMGL with the state-of-the-art methods in disease prediction tasks.
\vspace{-0.4cm}
\subsection{Datasets and Preprocessing}

\textbf{TADPOLE}~\cite{TADPOLE}: As a subset of the Alzheimer’s Disease Neuroimaging Initiative (ADNI) database (adni.loni.usc.edu), TADPOLE dataset contains features extracted from multi-modalites, which include MR, PET, cognitive tests, cerebro-spinal fluid (CSF) biomarkers, risk factors, clinical examinations and demographic information. For Alzheimer’s Disease prediction, we select 685 patients with 366-dimensional multi-modal features from TADPOLE, divided into 245 normal, 360 Mild Cognitive Impairment (MCI) and 80 Alzheimer’s Disease (AD) patients, respectively. Then, we divide the features according to the corresponding modalities and use the means of features for missing value filling.

\noindent\textbf{ABIDE}~\cite{ABIDE_data}: The Autism Brain Imaging Data Exchange (ABIDE) collected 1000 resting-state functional magnetic resonance imaging (R-fMRI) data with corresponding phenotypic data from 20 different sites. For Autism disease prediction, we select 871 patients, which are divided into 468 normal and 403 Autism Spectrum Disorder (ASD) patients. Then, for a fair comparison, we follow the preprocessing step as in ~\cite{popGCN}.
\vspace{-0.3cm}
\subsubsection{Implementation details} 
Without loss of generality, the standard GCN~\cite{GCN} is adopted as the prediction module, which can also be replaced by other GCNs.
 To stabilize the training process, we use a modular iterative training strategy, where a complete training epoch is performed by jointly training MaFF and AGL once, and then jointly training AGL and GCN once. The model uses Adam~\cite{Adam} as the optimizer and is implemented on the PyTorch platform. For hyper-parameters tuning, we set 4 attention heads of MaFF, and the other hyper-parameters are tuned through hyperopt~\cite{hyperopt}. 
\vspace{-0.3cm}
\subsection{Performance Comparisons}

\subsubsection{Baselines.} To evaluate the performance of MMGL, we choose to compare with several baselines, especially those that have achieved the state-of-the-art results in disease prediction tasks recently. Both PopGCN~\cite{popGCN} and InceptionGCN~\cite{InceptionGCN} are single-graph-based methods and two of the earliest works to use GCNs for disease prediction tasks. Multi-GCN~\cite{selfGCN} is a multi-graph-based method. In addition, we also compared with LGL~\cite{LGL} and EV-GCN~\cite{huang} which are the most related state-of-the-art works in disease prediction tasks.
\renewcommand{\thefootnote}{*}
\begin{table}[]

	\centering
	\vspace{-0.5cm}
	\caption{ Quantitative comparisons over two benchmark datasets.$(\%)$}
	\begin{tabular}{l|c c|c c}
		\hline
		\multicolumn{1}{c|}{\multirow{2}{*}{Methods}} & \multicolumn{2}{c|}{TADPOLE}                             & \multicolumn{2}{c}{ABIDE}                         \\ \cline{2-5} 
		\multicolumn{1}{c|}{}                  & \multicolumn{1}{c}{~ACC$(\%)$~} & ~AUC$(\%)$~  & \multicolumn{1}{c}{~ACC$(\%)$~} & ~AUC$(\%)$~  \\ \hline
		PopGCN~\cite{popGCN}                   & ~~$82.11\pm5.45$~~       & ~~$80.32\pm4.90$~~   & ~~$69.80\pm3.35$~~        & ~~$80.32\pm3.90$~~       \\
		InceptionGCN~\cite{InceptionGCN}       & ~~$74.39\pm1.24$~~       & ~~$81.17\pm1.60$~~   & ~~$72.69\pm2.37$~~        & ~~$72.81\pm1.94$~~       \\
		Multi-GCN\footnotemark[1]~\cite{selfGCN}
		& ~~$84.11\pm5.22$~~       & ~~$89.34\pm5.38$~~   & ~~$69.24\pm5.90$~~        & ~~$70.04\pm4.22$~~      \\
		EV-GCN~\cite{huang}                    & ~~$86.90\pm2.37$~~       & ~~$90.71\pm2.55$~~   & ~~{\color{blue}$85.90\pm4.47$}~~        & ~~{\color{blue}$84.72\pm4.27$}~~       \\
		LGL\footnotemark[1]~\cite{LGL}                         
		& ~~{\color{blue}$91.14\pm1.77$}~~ & ~~{\color{red}\textbf{93.92$\pm$1.61}}~~  & ~~$84.69\pm4.19$~~ & ~~$84.46\pm4.33$~~       \\ \hline
		MMGL                                   & ~~{\color{red}\textbf{93.73$\pm$1.70}}~~ & ~~{\color{blue}$93.81\pm1.80$}~~   & ~~{\color{red}\textbf{86.95$\pm$3.88}}~~        & ~~{\color{red}\textbf{86.74$\pm$3.77}}~~       \\ \hline
	\end{tabular}
	\vspace{-1cm}
	
	\label{results}

\end{table}
\footnotetext[1]{Due to the source code is not publicly available, the reported results here are our re-implementation of the original algorithms.}
\vspace{-0.2cm}
\subsubsection{Quantitative Results.}
We evaluate MMGL and other baselines on both the TADPOLE and ABIDE datasets using 10-fold stratified cross validation strategy, and the mean
scores and standard errors of Area Under Curve (AUC) and accuracy are reported. As shown in Table.~\ref{results},  it can be concluded that,
\textbf{(i)} compared to single-graph-based methods which are simply using meta-features and imaging features, the further use of multi-modal features can effectively improve the performance of the model; \textbf{(ii)} our MMGL outperforms Multi-GCN~\cite{selfGCN} by about 10.0\% and 17.5\% on TADPOLE and ABIDE datasets respectively, which just verifies that the effectiveness of MaFF to integration of multi-modal features; \textbf{(iii)} the approximate
2.2\% improvement of our MMGL on ABIDE dataset over LGL that is the state-of-the-art method is achieved, which demonstrates that our AGL may more effective compared to the graph learning in LGL.
	\begin{figure}[h]
	\centering
	\vspace{-0.3cm}
	\begin{tabular}{cccc}
		\includegraphics[scale=0.1]{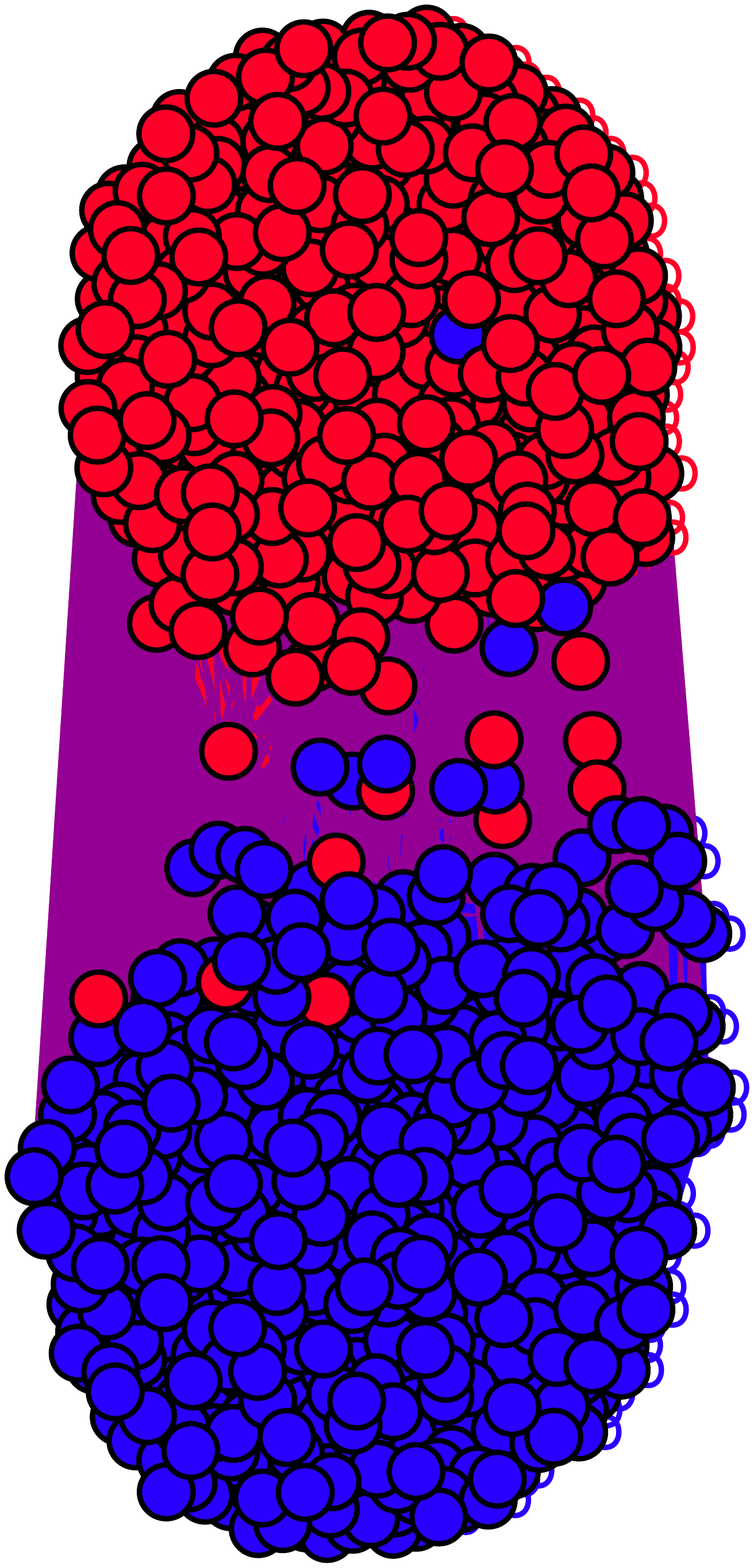} & 
		\includegraphics[scale=0.1]{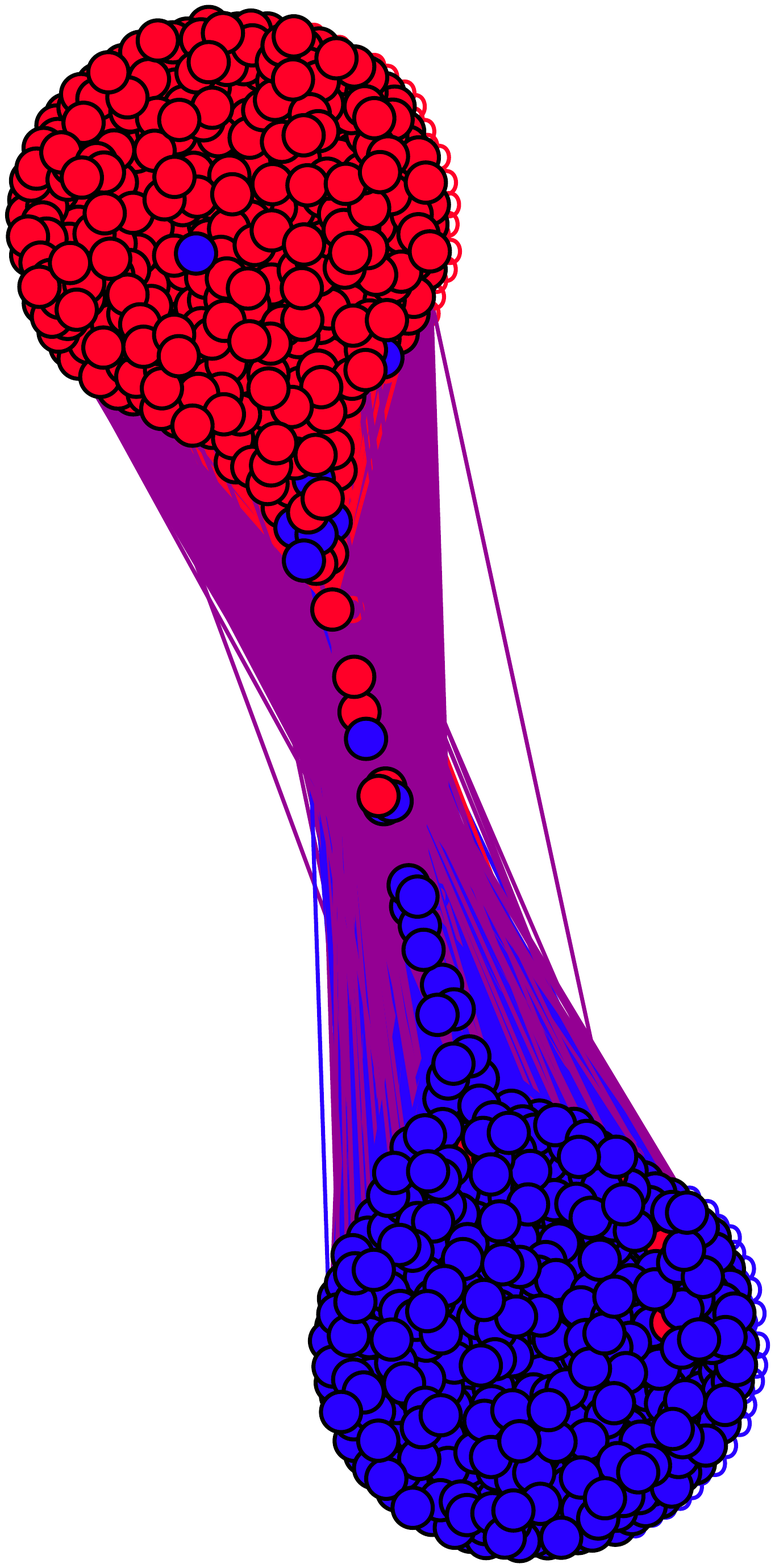} &
		\includegraphics[scale=0.1]{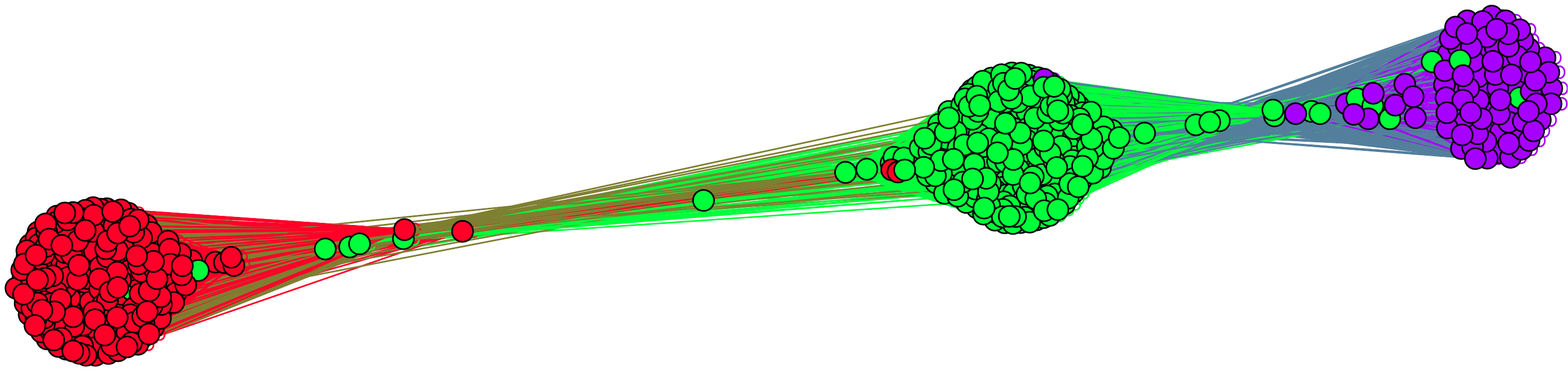} & 
		\includegraphics[scale=0.1]{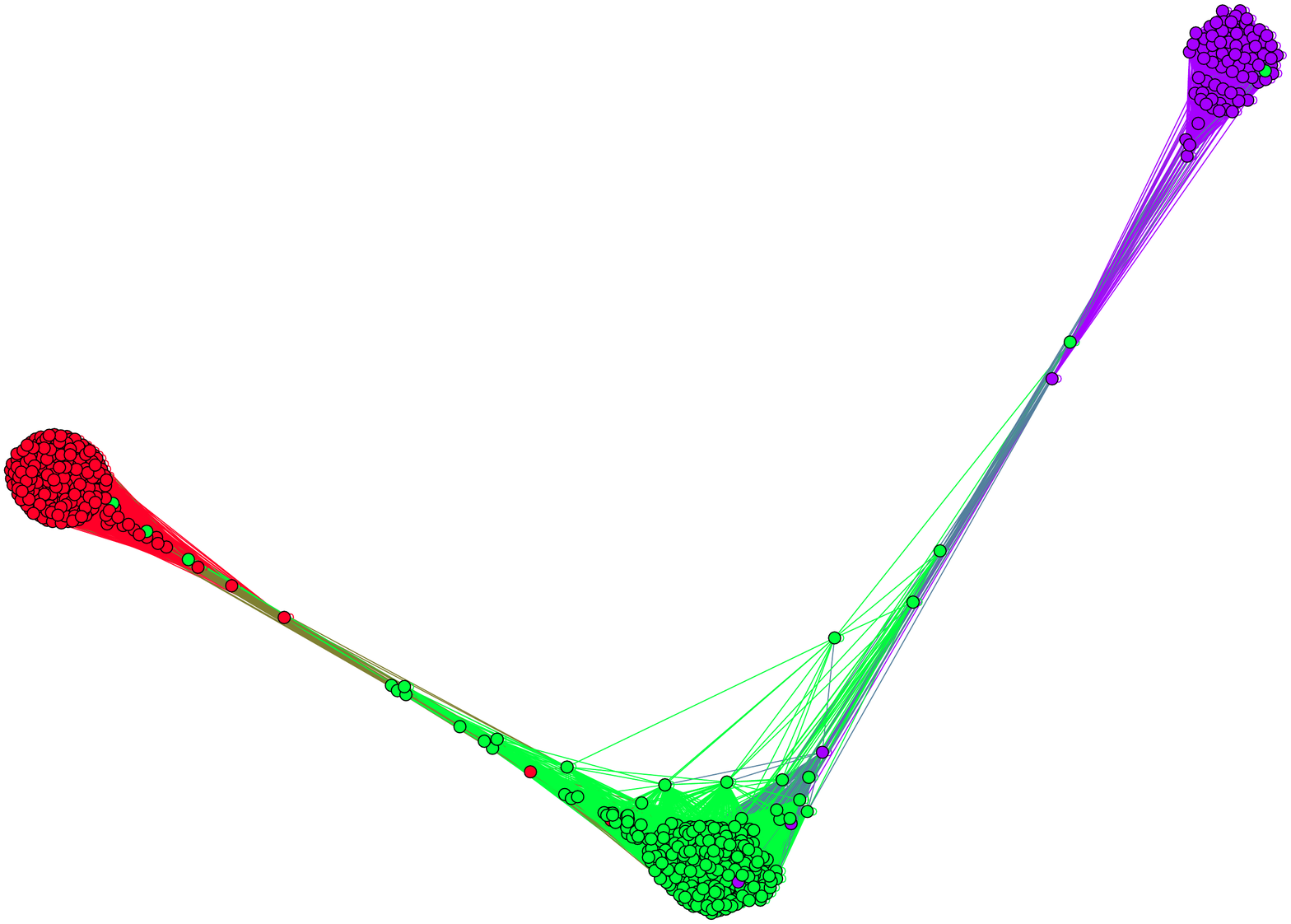}  \\
		(a) & (b)  & (c)   & (d)  
	\end{tabular}
	\vspace{-0.3cm}
	\caption{Visualization of the kNN graph and the learned graph on ABIDE and TADPOLE. Colors indicate different
		node labels. (a) kNN graph of ABIDE; (b) learned graph of ABIDE; (c) kNN graph of TADPOLE; (d) learned graph of TADPOLE.}
	\vspace{-0.3cm}
	\label{visualization} 
\end{figure}
\vspace{-0.3cm}
\subsubsection{Qualitative Results.} We visualize the kNN graphs and graph structures learned by MMGL, where the kNN graphs are constructed by the fused features. As illustrated in Fig.~\ref{visualization}, the kNN graph and the learned graph have overall similarity and form subgraphs corresponding to the patient classes, reflecting the differences between classes, which demonstrates the effectiveness of MaFF. Furthermore, compared to kNN graph, graph structure learned by AGL is more sparse between different classes of patients, indicating that AGL is better able to learn intra-class similarity while capturing inter-class differences. Thus, although the groundtruth graph does not exist, we can still see the superiority of AGL compared to kNN graph construction.
\vspace{-0.3cm}
\subsection{Ablation study}
\vspace{-0.2cm}

To validate the effectiveness of modal-attentional feature fusion (MaFF) and adaptive graph learning mechanism (AGL), we replace MaFF with MLP and concatenation operation respectively, and replace AGL with kNN graph construction based on RBF kernel and construction method of popGCN respectively. Table.~\ref{ablation} shows the ablation study results on different modules in our models. Specifically, the performance of the constructed graph of popGCN is the worst, especially on the ABIDE dataset, indicating that hand-constructed graphs are indeed not a desirable choice. Besides, AGL achieves favorable performance despite the absence of MaFF, which again validates the effectiveness of adaptive graph learning. More importantly, it can be observed that the combination of MaFF and AGL achieves a performance that far exceeds other combinations.

\begin{table}[]
	\centering
	\vspace{-0.2cm}
	\caption{ Quantitative evaluation of ablation studies on the TADPOLE and ABIDE datasets.}
	\begin{tabular}{l|cc|cc}
		\hline
		\multirow{2}{*}{} & \multicolumn{2}{c|}{TADPOLE} & \multicolumn{2}{c}{ABIDE} \\ \cline{2-5} 
		& ~ACC$(\%)$~                 & ~AUC$(\%)$~            & ~ACC$(\%)$~            & ~AUC$(\%)$~        \\ \hline
		MMGL              & ~~{\color{red}\textbf{93.73$\pm$1.70}}~~       & ~~{\color{red}\textbf{93.81$\pm$1.80}}~~   & ~~{\color{red}\textbf{86.95$\pm$3.88}}~~        & ~~{\color{red}\textbf{86.74$\pm$3.77}}~~           \\ \hline
		MLP+AGL           & ~~{\color{blue}89.42$\pm$2.22}~~      & ~~{\color{blue}90.30$\pm$2.34}~~ & ~~{\color{blue}85.07$\pm$4.68}~~ & ~~83.36$\pm$4.64~~           \\
		Concat+AGL        & ~~85.97$\pm$2.06~~      & ~~88.17$\pm$2.48~~ & ~~83.70$\pm$2.64~~ & ~~83.70$\pm$2.68~~           \\ \hline
		MaFF+popGCN       & ~~85.39$\pm$3.37~~      & ~~87.91$\pm$2.78~~ & ~~69.27$\pm$4.45~~ & ~~69.30$\pm$4.33~~           \\
		MaFF+kNN          & ~~87.97$\pm$3.69~~      & ~~89.67$\pm$2.65~~ & ~~84.19$\pm$3.83~~ & ~~{\color{blue}83.91$\pm$4.07}~~           \\ \hline
	\end{tabular}
	\vspace{-0.4cm}
	\label{ablation}
\end{table}
\vspace{-0.5cm}
\section{Conclusion}

In this paper, we propose a novel multi-modal graph learning framework named MMGL for disease prediction. For better integration of complementary information of multi-modalities, we propose a modal-attentional feature fusion module (MaFF) to achieve feature fusion considering inter-modal correlations. Furthermore, based on fused features, a lightweight adaptive graph learning mechanism is proposed to reveal the intrinsic connections among samples, constructing the optimal graph structure for downstream tasks. We have carried out extensive experiments on two disease prediction problems, and the results demonstrate the obvious superiority of our MMGL over currently available alternatives. More importantly, as a high modular inductive framework, MMGL provides a baseline that different variants of MMGL can be easily implemented to perform scenario-specific multimodal adaptive graph learning. Our ongoing research work will extend our MMGL to adaptive unified graph learning for more biomedical tasks.

%
%
%

\bibliographystyle{splncs04}
\bibliography{reference}
%

\end{document}